\documentclass[conference, a4paper]{IEEEtran}
\IEEEoverridecommandlockouts

	\usepackage[turkish]{babel}
	\usepackage[utf8]{inputenc} 
	\usepackage[T1]{fontenc}

        \usepackage{bbding}
        \usepackage{url}

\usepackage{cite}

\ifCLASSINFOpdf
  \usepackage[pdftex]{graphicx}
\else
\fi
\usepackage[cmex10]{amsmath}
\usepackage{multirow}
\usepackage{array}
\usepackage[lofdepth,lotdepth]{subfig}

\AtBeginDocument{%
  \renewcommand\tablename{TABLO}
}

\AtBeginDocument{%
  \renewcommand\abstractname{Abstract}
}

\usepackage{cite}
\usepackage{amsmath,amssymb,amsfonts}
\usepackage{algorithmic}
\usepackage{graphicx}
\usepackage{textcomp}
\usepackage{xcolor}
\def\BibTeX{{\rm B\kern-.05em{\sc i\kern-.025em b}\kern-.08em
    T\kern-.1667em\lower.7ex\hbox{E}\kern-.125emX}}

\usepackage{graphicx}
\usepackage{tabularx}
\usepackage{longtable}
\usepackage{adjustbox}
\usepackage{amsmath}
\usepackage{threeparttable}
\usepackage{makecell}
\usepackage{booktabs} 

\usepackage[turkish]{babel}

\usepackage{bbding}

\AtBeginDocument{%
  \renewcommand\tablename{TABLO}
}

\AtBeginDocument{%
  \renewcommand\abstractname{Abstract}
}
\usepackage{fancyhdr}

\fancypagestyle{firstpage}{%
 \fancyhf{}%
 \fancyhead[L]{8th International Artificial Intelligence and Data Processing Symposium (IDAP'24), \textit{Sept 21-22, 2024, Malatya, Türkiye}}
  
 \fancyfoot[L]{
  \normalsize{979-8-3315-3149-2/24/\$31.00 ~\copyright2024 IEEE}
 }%
}

\begin{document}


%
\title{Cosmos-LLaVA: Görselle Sohbet Etmek \\ Cosmos-LLaVA: Chatting with the Visual  
}



\author{
\IEEEauthorblockN{Ahmed Zeer}
\IEEEauthorblockA{\textit{Department of Computer Engineering} \\
\textit{Yildiz Technical University}\\
Istanbul, Turkey \\
ahmed.zeer@std.yildiz.edu.tr}
\and
\IEEEauthorblockN{Eren Dogan}
\IEEEauthorblockA{\textit{Department of Computer Engineering} \\
\textit{Yildiz Technical University}\\
Istanbul, Turkey \\
eren.dogan2@std.yildiz.edu.tr}
\and
\IEEEauthorblockN{Yusuf Erdem}
\IEEEauthorblockA{\textit{Department of Computer Engineering} \\
\textit{Yildiz Technical University}\\
Istanbul, Turkey \\
erdem.erdem@std.yildiz.edu.tr}
\and
\IEEEauthorblockN{Elif İnce}
\IEEEauthorblockA{\textit{Department of Computer Engineering} \\
\textit{Yildiz Technical University}\\
Istanbul, Turkey \\
elif.ince@std.yildiz.edu.tr}
\and
\IEEEauthorblockN{Osama Shbib}
\IEEEauthorblockA{\textit{Department of Computer Engineering} \\
\textit{Yildiz Technical University}\\
Istanbul, Turkey \\
osama.shbib@std.yildiz.edu.tr}
\and
\IEEEauthorblockN{M. Egemen Uzun}
\IEEEauthorblockA{\textit{Department of Computer Engineering} \\
\textit{Yildiz Technical University}\\
Istanbul, Turkey \\
egemen.uzun@std.yildiz.edu.tr}
\and
\IEEEauthorblockN{Atahan Uz}
\IEEEauthorblockA{\textit{Department of Computer Engineering} \\
\textit{Yildiz Technical University}\\
Istanbul, Turkey\\
atahan.uz@std.yildiz.edu.tr}
\and
\IEEEauthorblockN{M. Kaan Yuce}
\IEEEauthorblockA{\textit{Department of Computer Engineering} \\
\textit{Yildiz Technical University}\\
Istanbul, Turkey \\
kaan.yuce@yildiz.edu.tr}
\and
\IEEEauthorblockN{H. Toprak Kesgin}
\IEEEauthorblockA{\textit{Department of Computer Engineering} \\
\textit{Yildiz Technical University}\\
Istanbul, Turkey
 \\
tkesgin@yildiz.edu.tr}
\and
\IEEEauthorblockN{M. Fatih Amasyali}
\IEEEauthorblockA{\textit{Department of Computer Engineering} \\
\textit{Yildiz Technical University}\\
Istanbul, Turkey \\
amasyali@yildiz.edu.tr}
}



\maketitle

\thispagestyle{firstpage}

\begin{ozet}
Bu çalışmada bir Türkçe görsel talimat modeli geliştirilerek bu modelin performansını artırmaya yönelik çeşitli model mimarileri ve veri kümesi kombinasyonları derinlemesine incelenmiştir. Farklı büyük dil modelleri ve görüntü kodlayıcılarının bir araya getirilmesiyle oluşturulan Cosmos-LLaVA modeli, Türkçe dilindeki eksiklikleri gidermeye yönelik olarak tasarlanmıştır. Yapılan deneylerde, çeşitli veri kümeleri ile yapılan ince ayarların model performansını nasıl etkilediği detaylı olarak ele alınmıştır. Sonuçlar, model mimarisi ve veri kümesi seçiminin performans üzerinde önemli bir etkiye sahip olduğunu göstermektedir.
\end{ozet}
\begin{IEEEanahtar}
yapay zeka, doğal dil işleme, multimodelite, büyük dil modelleri, görselden soru cevaplama, üretici modeller, insan değerlendirmesi, yapay zeka hakemi
\end{IEEEanahtar}

\begin{abstract}
In this study, a Turkish visual instruction model was developed and various model architectures and dataset combinations were analysed to improve the performance of this model. The Cosmos-LLaVA model, which is built by combining different large language models and image coders, is designed to overcome the deficiencies in the Turkish language. In the experiments, the effects of fine-tuning with various datasets on the model performance are analysed in detail. The results show that model architecture and dataset selection have a significant impact on performance.

\end{abstract}


\section{Giriş}

Günümüzde Doğal Dil İşleme alanında yapılan çalışmalar Büyük Dil Modellerinin (BDM) gelecekteki önemini ortaya koymaktadır.
Bu modellerin geniş kullanım alanları, yetkinliklerinin yalnızca metin işleme becerisiyle sınırlı kalmaması gerektiğini göstermektedir. 
Yazılı ifadelerin yanında bir resmin anlamlandırılması ve bu doğrultuda yanıtlar üretilebilmesi de büyük önem taşımaktadır. Diğer dillerde yapılan çalışmalarla kıyaslandığında yalnızca Türkçe için olan çalışmaların azlığı görülmektedir. Çalışmanın amacı, literatürdeki bu eksikliği gidermek ve sistematik bir yol izleyerek sonraki araştırmacılar için güçlü bir temel oluşturmaktır.

Bu bağlamda geçmiş çalışmalarda kullanılan veri kümeleri ile çalışmada oluşturulan veri kümeleri karşılaştırılmıştır. Seçilen veri kümeleri ve model mimarilerinin oluşturabileceği bütün kombinasyonlar test edilmiş, böylece en iyi performansın elde edilmesi amaçlanmıştır. Yapılan değerlendirmelerde ulaşılan bulgular, eğitim süreçlerindeki hiperparametre optimizasyonu gibi detaylarla birlikte gelecekte çalışmalar için yol gösterici niteliktedir.

\section{Mimari}

Bir dil modeline görüntüleri anlama ve yorumlama yeteneği kazandırmak amacıyla yapılan başarılı çalışmalarda\cite{liu2023llava} aşağıdaki yaklaşımlar izlenmiştir:

Modelin yapısında üç temel bileşen bulunmaktadır:
\begin{itemize}
 \item \textbf{Görüntü Kodlayıcısı}
  \item \textbf{Projeksiyon Matrisi}
    \item \textbf{Büyük Dil Modeli}

\end{itemize}

Görüntü kodlayıcısının çıktılarının herhangi bir BDM'ye aktarılabilmesi için belirli bir projeksiyon matrisine ihtiyaç duyulmaktadır. Modellerin ön eğitim süreçlerinde yalnızca bu projeksiyon matrisi eğitilmektedir. Bu işlemde projeksiyon matrisi, kodlayıcının çıktılarının
en iyi şekilde yorumlanmak üzere BDM'ye aktarılmasından sorumludur.

\textit{Fine-tuning} (ince ayar) aşamasına geçildiğinde ise hem dil modelinin hem de görüntü kodlayıcısının ağırlıkları eğitilmeye başlanmaktadır.
Bu aşamada, modelin sohbet yetenekleri geliştirilmekte ve çeşitli veri kümeleriyle eğitilerek performansı artırılmaktadır. Bu çalışmada Türkçede yüksek performans gösterdiği saptanan, farklı mimarilerden iki dil modeli kullanılmıştır: \textit{CosmosLLaMA-Instruct}\cite{cosmosLLaMa-Model-Card} ve \textit{TrendyolMistral}\cite{Trendyol-Model-Card}. Her iki deneyde de görüntü kodlayıcısı olarak geçmiş çalışmalarda\cite{liu2023llava} başarılı sonuçlar elde edilen \textit{openai/clip-vit-large}\cite{radford2021learningtransferablevisualmodels} modeli tercih edilmiştir.

\section{Veri Kümeleri}

Geçmiş çalışmalarda geliştirilen veri kümeleri,
açık kaynak bir model\cite{tiedemann-thottingal-2020-opus} aracılığıyla İngilizce'den Türkçeye çevrilmiştir. Modele verilen görseldeki metin ile çeviri sonucunda oluşan metnin birebir aynı olmaması sebebiyle bu veri kümesinin geniş ölçekli olarak kabul edilemeyeceği görülmüştür. Dolayısıyla Optik Karakter Tanıma (OCR) verilerinin İngilizce'den Türkçeye çevrilmesinin anlamlı olmayacağı sonucuna varılmıştır. 
Bu sorunu çözmek için geliştirilen yaklaşımlar, aşağıdaki iki alt başlıkta ayrıntılı olarak incelenmektedir. Çalışmanın kalanında önceden oluşturulmuş veri kümesi \textbf{99eren} \cite{99eren}, bu çalışmada geliştirilen veri kümesi ise \textbf{CosmosVQA} \cite{llava-pretrain, llava-finetune, turkce-kitap} şeklinde ifade edilmiştir. Veri kümelerinin alt indisinin \textit{F} olması ince-ayar kümesinden, \textit{P} olması ise  ön eğitim kümesinden bahsedildiğini gösterecektir. 

\begin{equation}
\text{CosmosVQA} = \text{CosmosVQA}_P \cup \text{CosmosVQA}_F
\end{equation}

\begin{equation}
\text{99eren} = \text{99eren}_P \cup \text{99eren}_F
\end{equation}

\subsection{$\text{CosmosVQA}_P$ \cite{llava-pretrain}} 3 Milyon resim ve açıklama içeren açık kaynak CC3M\cite{inproceedings} veri kümesinden alınan 660 bin veri, yapılan değerlendirmelerde en iyi sonucu verdiği görülen çeviri aracı DeepL\cite{DeepL} kullanılarak Türkçeye çevrilmiştir.
Tablo \ref{table:training_data_samples}'de gösterildiği gibi bu aşamadaki verilerin amacı modele görüntüyü anlamlandırma yeteneğini kazandırmaktır.

\subsection{$\text{CosmosVQA}_F$ \cite{llava-finetune}} Aşağıda listelenen çeşitli kaynaklardan ve geniş bir görev yelpazesine sahip veri kümelerinden oluşmaktadır. Kitaplar kullanılarak oluşturulan veri kümesi dışındaki verilerin kaynakları İngilizcedir. Bu veriler ön eğitim verisi için kullanılan yöntemler ile Türkçeye çevrilmiştir.

\textbf{Türkçe-Kitap \cite{turkce-kitap}: } Tamamen Türkçe kaynaklardan toplanan 100 bin kitabın derlenmesiyle oluşturulmuş bir veri kümesidir. Veri setinin amacı, modelin OCR becerilerini geliştirerek görselde bulunan herhangi bir yazının model tarafından algılanmasını ve yorumlanmasını sağlamaktır.

\textbf{GQA \cite{hudson2018gqa}:} Modelin çoklu muhakeme becerileri, mekansal anlayış ve çok adımlı çıkarımlarını geliştirmeyi amaçlayan bu veri kümesinden 72 bin görsel içeren bir veri seti oluşturulmuştur.  

\textbf{COCO \cite{lin2015microsoftcococommonobjects}:} Nesne tespiti ve segmentasyon görevlerini içeren bu veri kümesinden 364 bin görsel içeren bir veri seti oluşturulmuştur. 

\textbf{VG \cite{krishna2016visualgenomeconnectinglanguage}:} Nesneler arası ilişkilerin çözümlenmesini ele alan bu veri kümesinden 86 bin görsel içeren bir veri seti oluşturulmuştur.

Her veri setine ait örnekler Tablo \ref{table:training_data_samples}'de gösterilmiştir. Bütün veri kümeleri açık kaynak olarak paylaşılmıştır. \cite{llava-pretrain, llava-finetune, turkce-kitap}


\begin{table*}[h]
\centering
\caption{İnce ayar veri kümesinden örnekler}
\begin{adjustbox}{max width=\textwidth}
\begin{tabularx}{\textwidth}{|X|X|X|X|X|}
\hline
\textbf{GQA} & \textbf{CosmosVQA} & \textbf{VG} & \textbf{COCO} & \textbf{CC3M} \\ \hline

\textbf{Kullanıcı:} Gökyüzü karanlık mı? \newline
\textbf{Model:} Evet \newline
\textbf{Kullanıcı:} Sokakta ne var? \newline
\textbf{Model:} Araba \newline
\textbf{Kullanıcı:} Caddede otobüs var mı? \newline
\textbf{Model:} Hayır \newline
\textbf{Kullanıcı:} Trafik sinyali sağda mı yoksa solda mı? \newline
\textbf{Model:} Sol \newline \newline
\includegraphics[width=0.18\textwidth]{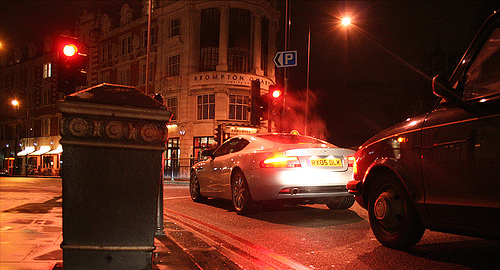}
&
\textbf{Kullanıcı:} Kitabın ismini öğrenebilir miyim? \newline
\textbf{Model:} Masumiyet Müzesi \newline
\textbf{Kullanıcı:} Bu kitap hangi türde bir kitaptır? \newline
\textbf{Model:} Aşk romanı \newline
\textbf{Kullanıcı:} Bu kitabın yazarı kimdir? \newline
\textbf{Model:} Orhan Pamuk \newline \newline
\includegraphics[width=0.15\textwidth]{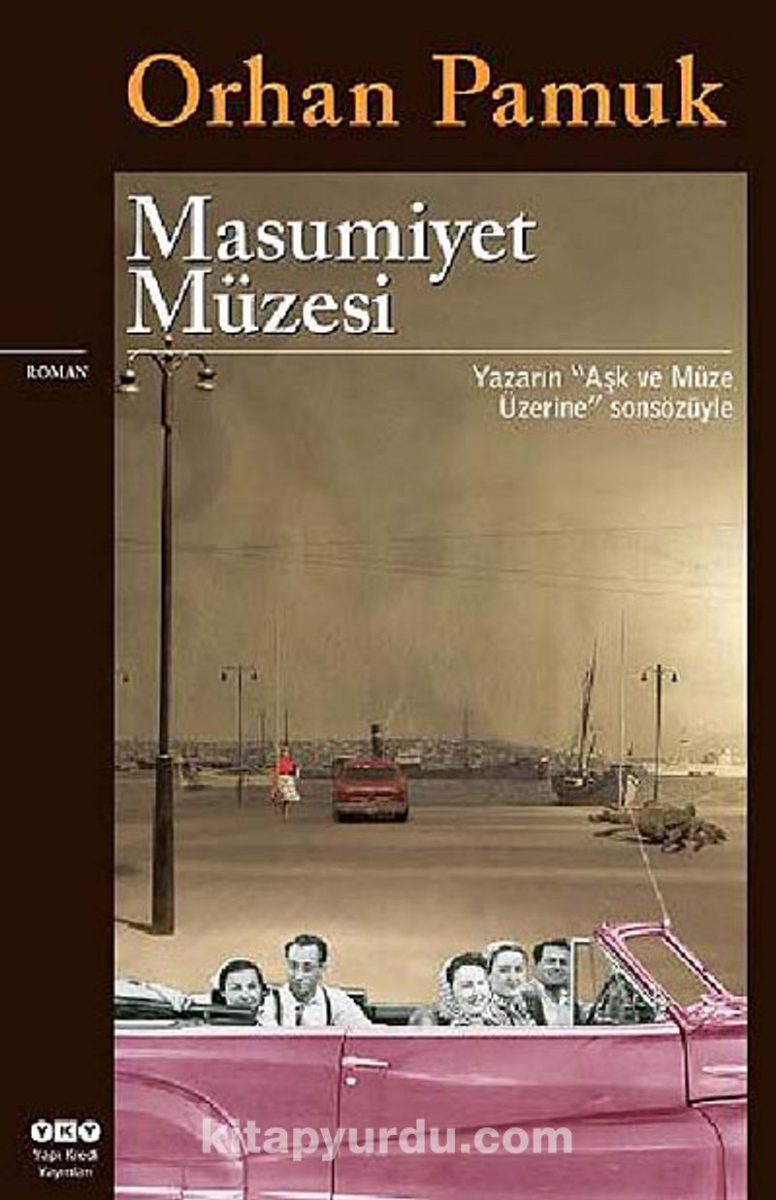}
&
\textbf{Kullanıcı:} Lütfen bu bölge için kısa bir açıklama yapınız: [0.69, 0.35, 0.79, 0.69]. \newline
\textbf{Model:} Bu bir insan. \newline
\textbf{Kullanıcı:} Lütfen bu cümlenin tanımladığı bölgenin sınırlayıcı kutu koordinatını verin: tenis kortundaki beyaz çizgi. \newline
\textbf{Model:} [0.0, 0.7, 0.9, 0.82] \newline \newline
\includegraphics[width=0.17\textwidth]{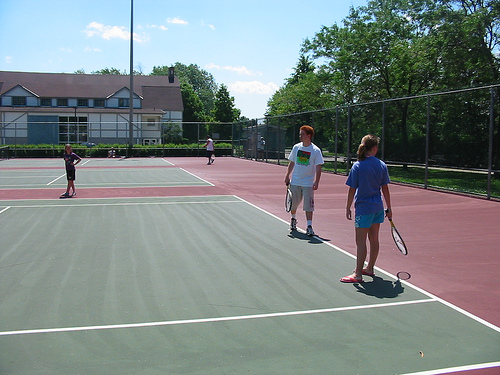}
&
\textbf{Kullanıcı:} Lütfen bu bölge için kısa bir açıklama yapınız: [0.3, 0.36, 0.62, 0.59]. \newline
\textbf{Model:} Fotoğrafçıya bakan bir bilgisayar ekranı. \newline
\textbf{Kullanıcı:} Lütfen bu cümlenin tanımladığı bölgenin sınırlayıcı kutu koordinatını verin: en sağdaki ekran. \newline
\textbf{Model:} [0.3, 0.36, 0.62, 0.59] \newline \newline
\includegraphics[width=0.15\textwidth]{./finetune_data_images/coco_000000449094.jpg}

&
\textbf{Kullanıcı:} Sunulan görüntüyü temsil eden kompakt bir anlatı oluşturun.
\newline
\textbf{Model:} çiftler için toplu nikah töreni düzenlendi. \newline \newline
\includegraphics[width=0.15\textwidth]{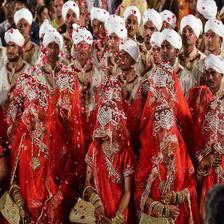}

\\ \hline 
\end{tabularx}
\end{adjustbox}
\newline

\label{table:training_data_samples}
\end{table*}

\section{Eğitim}

Çalışmada en iyi modele ulaşabilmek için birçok veri kümesi ve model mimarisi kombinasyonu denenmiştir. Bu bölümde genel olarak \textit{CosmosLLaMA-Instruct} modellerinin geçirdiği eğitim sürecinden bahsedilecektir.

\subsection{Ön eğitim}
Ön eğitim sürecinde yalnızca projeksiyon matrisi eğitilmektedir. Bu matris görüntü kodlayıcısının elde ettiği bilgileri dil modeline aktarmaktan sorumludur ve eğitimi aşağıda belirtilen ayarlarla 7 saat sürmektedir.  Eğitim 4 adet NVIDIA A100 GPU kullanılarak gerçekleştirilmiş olup her bir batch büyüklüğü 16 olacak şekilde ayarlanmıştır. Eğitimde \textit{learning rate scheduler} olarak kosinüs kullanılmış olup \textit{learning rate} için 1e-3 değeri belirlenmiştir. \textit{Gradient accumulation} değeri 4 olarak seçilmiş ve eğitim yalnızca 1 epoch üzerinden gerçekleştirilmiştir. Ayrıca \textit{warmup} oranı \%3 olarak ayarlanmıştır.



\begin{figure}[h]
	\centering
	\shorthandoff{=}  
	\includegraphics[scale=0.2]{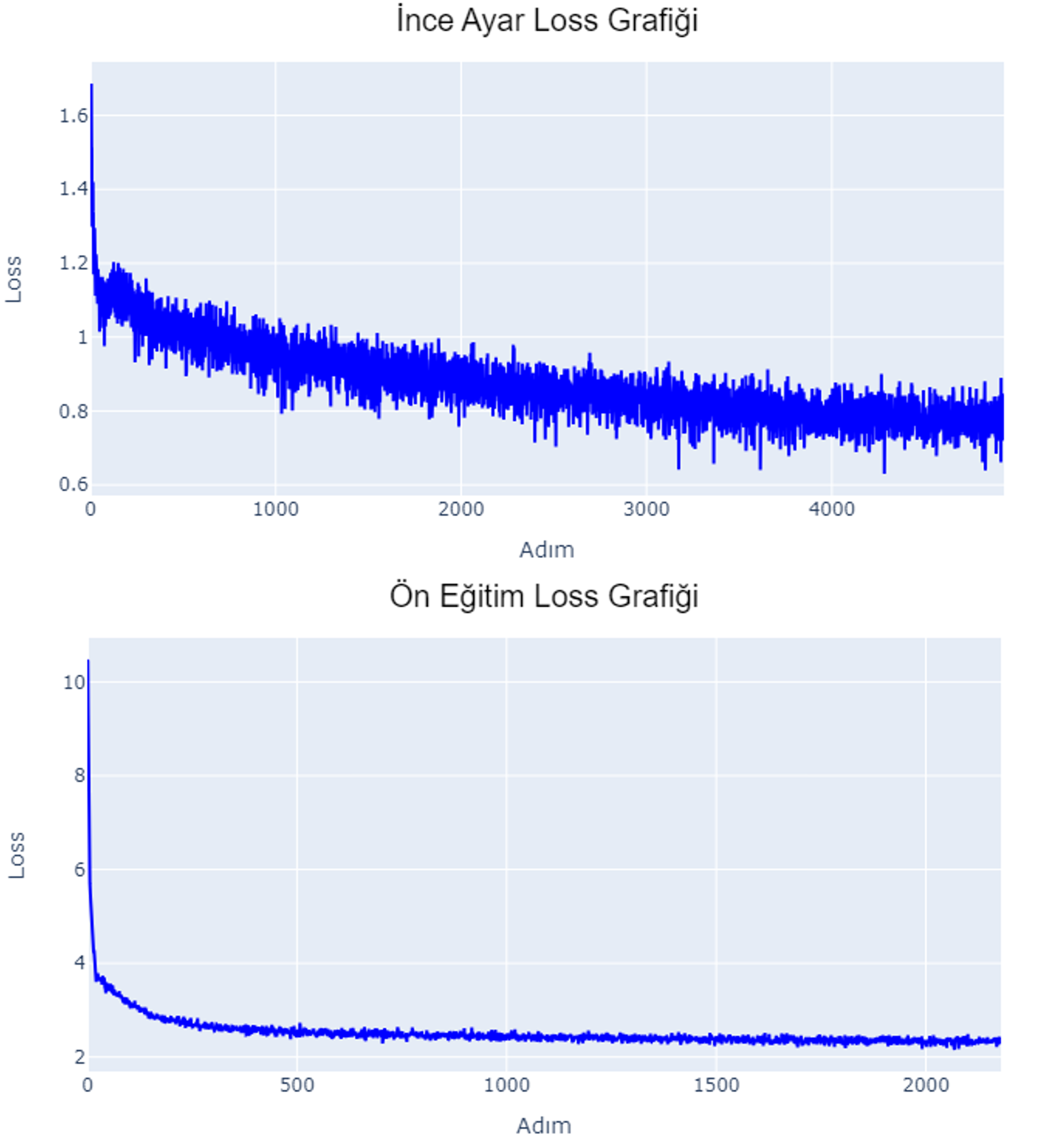}
	\shorthandon{=} 
	\caption{Eğitimlere Ait Loss Grafikleri}
	\label{losses_figure}
\end{figure}

\subsection{İnce-Ayar}

İnce-ayar aşamasında projeksiyon matrisi ve dil modeli aynı anda eğitilmektedir. Bu süreçte batch büyüklüğü 4 olarak belirlenmiş ve 4 adet NVIDIA A100 GPU kullanılarak eğitim gerçekleştirilmiştir. \textit{Learning rate} 2e-5 olarak ayarlanmış ve eğitim 1 epoch boyunca sürdürülmüştür. Eğitim sırasında, \textit{gradient accumulation} adımları 8 olarak belirlenmiş ve \textit{warmup} oranı \%3 olarak ayarlanmıştır. Eğitim toplam 20 saat sürmüştür.

İnce-ayar eğitimi ile modelin ağırlıkları, görüntülere dair talimatların etkin bir şekilde yerine getirilmesi için optimize edilmiştir.


\begin{figure}[h]
	\centering
	\shorthandoff{=}
	\includegraphics[scale=0.35]{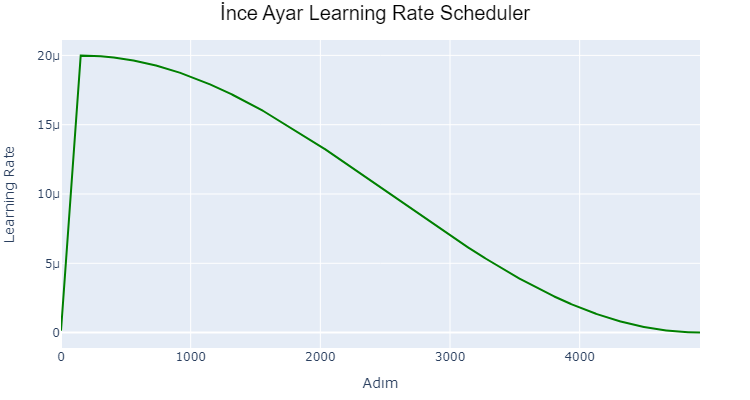}
	\shorthandon{=}
	\caption{İnce Ayar Eğitimde Learning Rate'in Değişimi}
	\label{lr_figure}
\end{figure}

\section{Değerlendirme Metodu}

Eğitimler sonucu elde edilen modelleri değerlendirmek için seçilen görsellere ait 100 adet soru ve cevaptan oluşan bir test kümesi oluşturulmuştur. Bu test kümesi ile 3 farklı metodoloji ile değerlendirme yapılmıştır.

\subsection{Hakem: Dil Modeli }
Bu çalışmanın yapıldığı Ağustos 2024 tarihinde çoğu ölçüme göre en iyi dil modeli olan OpenAi'ın GPT-4o\cite{OpenAI2024GPT4o} modeli hakem olarak kullanılmıştır.
Sorular, resimler ve modellerin verdikleri cevaplar hakem olarak belirlenen GPT-4o modeline verilerek modelin yanıta ne kadar doğru olduğuna göre 100 üzerinden bir puan vermesi istenmiştir. 
GPT-4o'ya puanlattırmak için kullanılan talimat formatı aşağıdaki gibidir:

\textbf{\{görsel\}}

\textbf{Soru:} \texttt{\{soru\}}

\textbf{Cevap:} \texttt{\{model-cevabı\}}

\textit{Bir görsel, görselle ilgili bir soru ve verilen cevap gösterilmiştir. Çok iyi kaliteli cevaplara 100, iyilere 80, ortalamalara 30, kötülere 0 puan ver. Sadece puanı tam sayı olarak yaz, hiçbir açıklama yapma.}

Ardından GPT-4o'nun verdiği yanıt ile modelin o soruya ne kadar iyi yanıt verebildiği 100 üzerinden bir puan ile ölçülmüştür.  GPT-4o modelinin yanıtları deterministik olmaması ve aynı cevaba farklı puanlar verilebilmesi nedeniyle her soru için 5 kez değerlendirme yapılarak bu değerlendirmelerde verilen puanların ortalaması alınmıştır.

Bu puanlama için kullanılan test kümesi 4 alt kümeden oluşmaktadır: OCR, Kompleks, Tanımlama ve Detay. Bu kümelerin içeriği aşağıda açıklanmıştır:

\begin{itemize}
    \item \textbf{OCR:} Toplam 10 kitap içeren bu veri kümesi, her kitap için gerçek hayatta çekilmiş bir adet fotoğraf ile bir dijital kapak görselini içermektedir. Veri kümesi, kitabın kapağında yer alan yazıların ne olduğunu sorgulayan 20 sorudan oluşmaktadır.
    
    \item \textbf{Kompleks:} Bu kategori, modelin görsel içeriklere dayanarak üst düzeyde akıl yürütme ve çıkarımlar yapmasını, görselin barındırdığı bağlamları anlamasını gerektiren soruları içerir.

    \item \textbf{Tanımlama:} Bu kategori, modelden basit ve doğrudan yanıtlar bekleyen, görsellerdeki temel öğeleri tanımlamayı veya açıklamayı amaçlayan sorulardan oluşur.

    \item \textbf{Detay:} Bu kategorideki sorular, modelin görsellerin kapsamlı ve ayrıntılı bir analizini yapmasını gerektirir. Görseldeki birden fazla öğeyi ve bunlar arasındaki ilişkileri yakalamayı hedefler.
\end{itemize}

\begin{table*}
  \centering
  \caption{GPT-4o Hakemliğinde Model Performanslarının Karşılaştırılması }
  \begin{threeparttable}
    \begin{tabular}{l@{\qquad}r@{\qquad}r@{\qquad}r@{\qquad}r@{\qquad}r}
      \textbf{Model} & \textbf{OCR} & \textbf{Kompleks} & \textbf{Tanımlama} & \textbf{Detay} & \textbf{* TOPLAM} \\ \midrule\midrule 
        Mistral + $\text{CosmosVQA}_P$ + $\text{99eren}_F$ & 87.5 & 54.1 & 55.8 & 67.9 & 61.3 \\ 
        \cmidrule(l r){1-6}

        Mistral + $\text{99eren}_P$ + $\text{99eren}_F$ & 85.5 & 91.2 & 85.8 & 68.4 & 81.9 \\ 
        \cmidrule(l r){1-6}

        LLaMA + $\text{99eren}_P$ + $\text{99eren}_F$ & 76.5 & 89.6 & 86.5 & 75.1 & 83.3 \\ 
        \cmidrule(l r){1-6}

        \textbf{LLaMA + $\text{CosmosVQA}_P$ + $\text{99eren}_F$} & 68.0 & \textbf{95.3} & \textbf{89.6} & \textbf{77.2} &  \textbf{85.9} \\ 
        \cmidrule(l r){1-6}

        Mistral + $\text{99eren}_P$ + $\text{CosmosVQA}_F$ & 93.8 & 91.4 & 86.9 & 61.1 & 81.0 \\ 
        \cmidrule(l r){1-6}

        Mistral + $\text{CosmosVQA}_P$ + $\text{CosmosVQA}_F$ & 93.3 & 90.1 & 85.2 & 69.0 & 82.4 \\ 
        \cmidrule(l r){1-6}

        LLaMA + $\text{CosmosVQA}_P$ + $\text{CosmosVQA}_F$ & 96.2 & 92.7 & 88.6 & 64.4 & 83.1 \\ 
        \cmidrule(l r){1-6}

        LLaMA + $\text{99eren}_P$ + $\text{CosmosVQA}_F$ & \textbf{97.7} & 91.5 & 82.3 &  74.3 & 84.0 \\ 
        \cmidrule(l r){1-6}
        
        GPT-4.0-mini & 96.2 & 97.9 & 97.0 & 97.4 & 96.8 \\ 
        \cmidrule(l r){1-6}
        GPT-4.0 & 100.0 & 99.0 & 99.0 & 99.5 & 99.2 \\ 
      \midrule\midrule
    \end{tabular}

    \begin{tablenotes}
      \item[*] OCR, Tanımlama, Kompleks ve Detay sütunları, modellerin farklı görevlerdeki başarısını yansıtır.
      \item[**] * TOPLAM puanı, bu görevlerdeki genel performansı temsil eder.
    \end{tablenotes}
  
  \end{threeparttable}

  \label{table:model_performances_table}
\end{table*}


\subsection{Hakem: İnsan Etiketleyiciler}
Modellerin arasındaki farkı ortaya koyabilecek güvenilir
yöntemlerden bir tanesi de model cevaplarını insan oylamasına
sunmaktır. Bu çalışmada insan oylaması sonuçlarının ölçümünde ELO skorları kullanılmıştır. ELO ölçütü, insan oylamasına dayalı olarak dil modellerinin performansını birbirleriyle karşılaştırmak için yaygın olarak kullanılmaktadır \cite{chiang2024chatbot}.

Çalışmada hakem olarak 10 kişi görev almış ve toplam 2000 oy kullanılmıştır. Etiketleyicilere bir görsel, görsele dair bir soru ve anonim halde iki farklı modelin soruya verdikleri yanıt sunularak 4 seçenekten birini seçmeleri istenmiştir:\\
- Model A daha iyi \\
- Model B daha iyi \\
- İkisi de iyi \\
- İkisi de kötü 

Oylamanın başlangıcında bütün modeller 1000 ELO puanına sahiptirler. Her karşılaştırmada tercih edilen model puan kazanırken diğer model puan kaybetmektedir. Yüksek ELO'lu bir modeli yenmek, düşük ELO'lu bir modeli yenmekten daha fazla puan kazandırmaktadır. Yüksek ELO puanları diğer modellere göre üstün performans sergilendiği anlamını taşımaktadır.

\begin{table}
  \centering
  \caption{İnsan Hakemliğinde Model Performanslarının Karşılaştırılması}
  \begin{threeparttable}
    \begin{tabular}{l@{\qquad}r}
      \textbf{Model} & \textbf{ELO Puanı} \\ \midrule\midrule 
        GPT-4.0 & 1320.01 \\ 
        \cmidrule(l r){1-2}
        GPT-4.0-mini & 1243.05 \\ 
        \cmidrule(l r){1-2}
        \textbf{Mistral + $\text{CosmosVQA}_P$ + $\text{CosmosVQA}_F$}  & \textbf{1082.54} \\ 
        \cmidrule(l r){1-2}
        LLaMA + $\text{99eren}_P$ + $\text{CosmosVQA}_F$  & 1058.10 \\ 
        \cmidrule(l r){1-2}
        Mistral + $\text{99eren}_P$ + $\text{CosmosVQA}_F$  & 1038.23 \\ 
        \cmidrule(l r){1-2}
        LLaMA + $\text{CosmosVQA}_P$ + $\text{CosmosVQA}_F$ & 1013.60 \\ 
        \cmidrule(l r){1-2}
        LLaMA + $\text{CosmosVQA}_P$ + $\text{99eren}_F$ & 1008.74 \\ 
        \cmidrule(l r){1-2}
        LLaMA + $\text{99eren}_P$ + $\text{99eren}_F$ & 946.31 \\ 
        \cmidrule(l r){1-2}
        Mistral + $\text{99eren}_P$ + $\text{99eren}_F$ & 850.56 \\ 
        \cmidrule(l r){1-2}
        Mistral + $\text{CosmosVQA}_P$ + $\text{99eren}_F$ & 566.22 \\ 
      \midrule\midrule
    \end{tabular}

    \begin{tablenotes}
      \item[*] Daha yüksek ELO puanları görevde üstün performansı temsil eder.
    \end{tablenotes}
  \end{threeparttable}
  \label{table:human_voting_results}
\end{table}

\subsection{İkili Sınıflandırma}
Basit sorulardan oluşan bu küme, modelin sınıflandırma yeteneklerini ölçmede kullanılmıştır. Örnek: "\textit{Bu resimde kedi var mı? Tek bir kelimeyle cevap ver.}"

Modellerin cevapları arasından "Evet." veya "Hayır." şeklinde olmayanlar kullanıcı isteğini yerine getirmediği için yanlış olarak kabul edilmiştir.  Modellere ilişkin sonuçlar Tablo \ref{table:model_metrics_table}'de yer almaktadır.

\begin{table*}
  \centering
  \caption{Sınıflandırmada Model Performanslarının Karşılaştırılması}
  \begin{threeparttable}
    \begin{tabular}{l@{\qquad}c@{\qquad}c@{\qquad}c@{\qquad}c}
      \textbf{Model} & \textbf{Accuracy} & \textbf{Precision} & \textbf{Recall} & \textbf{F1 Score} \\ \midrule\midrule 
        LLaMA + $\text{99eren}_P$ + $\text{99eren}_F$ & 0.61 & 0.57 & 0.93 & 0.71 \\ 
        \cmidrule(l r){1-5}
        LLaMA + $\text{99eren}_P$ + $\text{CosmosVQA}_F$ & 0.59 & 0.55 & 0.91 & 0.69 \\ 
        \cmidrule(l r){1-5}
        \textbf{LLaMA + $\text{CosmosVQA}_P$ + $\text{99eren}_F$} & \textbf{0.69} & \textbf{0.62} & \textbf{0.96} & \textbf{0.75} \\ 
        \cmidrule(l r){1-5}
        LLaMA + $\text{CosmosVQA}_P$ + $\text{CosmosVQA}_F$ & 0.58 & 0.55 & 0.89 & 0.68 \\ 
        \cmidrule(l r){1-5}
        Mistral + $\text{99eren}_P$ + $\text{99eren}_F$ & 0.56 & 0.53 & 0.89 & 0.67 \\ 
        \cmidrule(l r){1-5}
        Mistral + $\text{99eren}_P$ + $\text{CosmosVQA}_F$ & 0.60 & 0.56 & 0.91 & 0.69 \\ 
        \cmidrule(l r){1-5}
        Mistral + $\text{CosmosVQA}_P $  + $\text{99eren}_F$ & 0.00 & 0.00 & 0.00 & 0.00 \\ 
        \cmidrule(l r){1-5}
        Mistral + $\text{CosmosVQA}_P $ + $\text{CosmosVQA}_F$ & 0.60 & 0.56 & 0.91 & 0.69 \\ 
      \midrule\midrule
    \end{tabular}

    \begin{tablenotes}
      \item[*] Bu tablo, farklı modellerin çeşitli metrikler açısından değerlendirilmesini gösterir.
    \end{tablenotes}
  
  \end{threeparttable}
    \label{table:model_metrics_table}
\end{table*}

\begin{table*}[h]
\caption{Model Çıktı Örnekleri}
\begin{adjustbox}{max width=\textwidth}
\begin{tabularx}{\textwidth}{|X|X|}
\hline
\textbf{Örnek 1: Görsel Açıklaması Üretme} & \textbf{Örnek 2: Akıl Yürütme} \\ \hline

\includegraphics[width=0.45\textwidth]{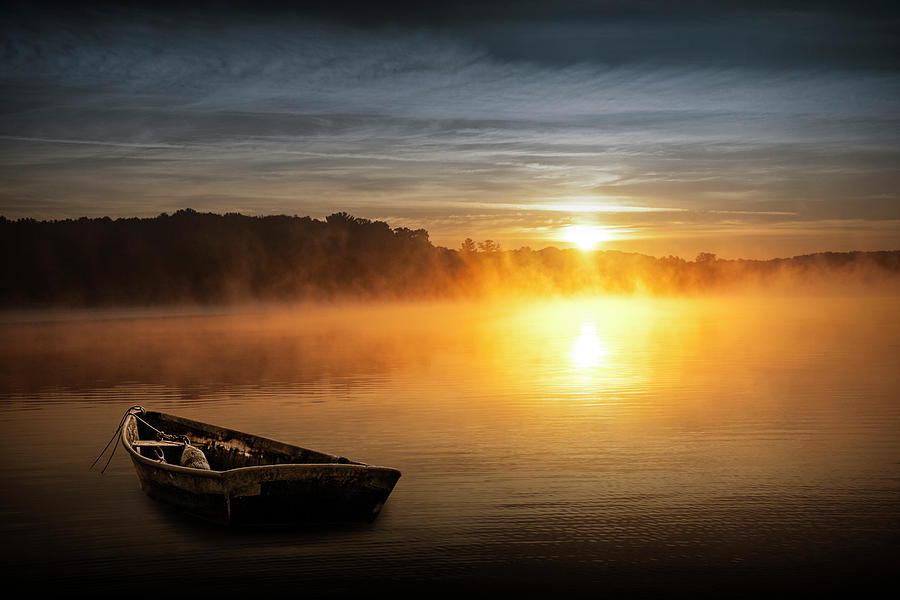} &
\includegraphics[width=0.45\textwidth]{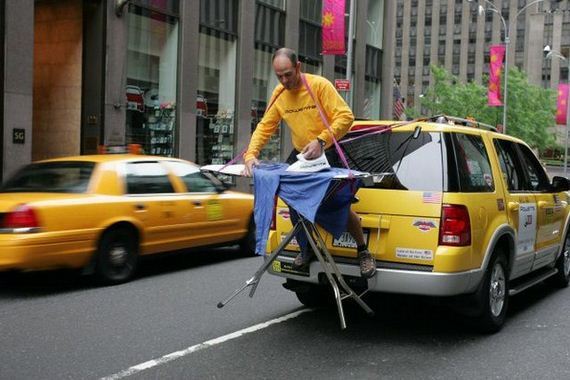} \\ 

\textbf{Kullanıcı:} Güneşin tamamen batmasının ardından bu görüntünün nasıl değişmesi beklenebilir? \newline
\textbf{Model:} Gün batımından sonra, güneşin ışığı yavaş yavaş azalacak ve suyun üzerindeki sıcak, altın rengi tonlar daha soğuk, daha koyu bir renge dönüşecektir. Bu değişim, sakin ve dingin bir atmosfer yaratarak sahnenin genel ambiyansını etkileyecektir. Ayrıca güneşin batışı nedeniyle gölün üzerindeki ışıklandırma değişebilir ve bu da suyun üzerinde yüzen teknenin görünürlüğünü etkileyebilir. &
\textbf{Kullanıcı:} Resimdeki kişinin tişörtü hangi renktir? \newline
\textbf{Model:} Resimdeki kişinin tişörtü sarıdır. \newline \newline
\textbf{Kullanıcı:} Ütü yapan adamla ilgili garip bir durum var mı? \newline
\textbf{Model:} Evet, adamın ütü masasını bir arabanın arkasına yerleştirdiği garip bir durum var. Bu alışılmadık bir durum çünkü ütü masaları genellikle iç mekanlarda, özellikle ütü masaları için tasarlanmış odalarda veya alanlarda bulunur. Adamın ütü masasını bir arabanın arkasına yerleştirmesi, ütü masasını kullanırken hareket halindeyken dengesini ve güvenliğini sağlamak için ekstra çaba ve dikkat gerektirebilir. \newline \\ 

\hline 
\end{tabularx}
\end{adjustbox}

\begin{tablenotes}
      \item[*] * Tabloda yer alan örnek çıktılar \textit{LLaMA + $\text{99eren}_P$ + $\text{CosmosVQA}_F$} modeli kullanılarak \newline elde edilmiştir. Model açık kaynak olarak paylaşılmıştır \cite{cosmosllava}.
\end{tablenotes}

\label{table:model_outputs}
\end{table*}

\section{Model Performanslarının Karşılaştırılması}
Çalışmanın başlangıcında açık kaynak dil modellerinin performanslarının değerlendirildiği liderlik tablolarında \cite{OpenLLMTurkishLeaderboard2024} birinci olan \textit{CosmosLLaMA-Instruct}'ın  bu başarısından dolayı çalışmada üstün performans göstermesi beklenmiştir. Ancak sonuçlar değerlendirildiğinde elde edilen modellerin performanslarının kullanılan dil modelinin mimarisine bağlı olarak değişiklik gösterdiği gözlemlenmiştir. Ayrıca DeepL kullanılarak çevirilen verilerin, açık kaynak bir çeviri modeline\cite{tiedemann-thottingal-2020-opus} göre, daha başarılı ve isabetli sonuçlar vermesi beklenirken diğer veri setleri ile aralarında büyük bir fark meydana gelmediği görülmüştür.

Modelin mimarisi ile kullanılan ön eğitim veri kümesine bağlı olarak en uygun ince-ayar kümesinin de değiştiği gözlemlenmiştir. Genel olarak bütün mimarilerde en iyi performansı gösteren bir veri kümesi elde edilememiştir. Eğitilen modellerin değerlendirme sonuçları incelendiğinde modellerin birbirlerinden farklı alanlarda iyi performans sergiledikleri ve bütün görevlerde mutlak başarıya sahip bir model olmadığı sonucuna varılmıştır. 

Eğitilen modellerin değerlendirme ölçütleri karşılaştırıldığında farklı ölçütlerde en iyi performansı gösteren modellerin benzersiz olduğu görülmüştür. 
GPT-4o hakemliğinde birinci model \textit{LLaMA + $\text{CosmosVQA}_P$ + $\text{99eren}_F$} olurken insan oylamasında ise birinci model  \textit{Mistral + $\text{CosmosVQA}_P$ + $\text{CosmosVQA}_F$} olmuştur. İkili sınıflandırma görevinde, \textit{LLaMA + $\text{CosmosVQA}_P$ + $\text{CosmosVQA}_F$} modeli en yüksek performansı sergilemiştir.



\subsection{GPT-4o Oylaması Analizi}

Kategori bazlı bir inceleme yapıldığında OCR görevinde özgün Türkçe içeriği sayesinde $\text{CosmosVQA}_F$ ile ince ayarı yapılan modellerin skorlarının 90'ın üstünde olduğu görülmüştür. Ancak en düşük genel skora sahip olan $\text{CosmosVQA}_P$ + $\text{99eren}_F$ modelinin de bu kategorideki başarısı göze çarpmaktadır. \textit{Detay} kategorisinde ise LLaMA modellerinin performansının diğer modellere göre yüksek olduğu görülmüştür. En yüksek genel skora sahip model olan \textit{LLaMA + $\text{CosmosVQA}_P$ + $\text{99eren}_F$}, aynı zamanda OCR kategorisinde en kötü performans gösteren model olmuştur. Genel sıralamada ikinci olan \textit{LLaMA + $\text{99eren}_P$ + $\text{CosmosVQA}_F$} modelinin OCR görevinde GPT4o-mini'yi\cite{OpenAI2024GPT4o-mini} geçtiği gözlemlenmiştir. Görevlere ait model skorları Tablo \ref{table:model_performances_table}'de yer almaktadır.

\subsection{İnsan Oylaması Analizi}

İnce ayar eğitimi $\text{CosmosVQA}_F$ ile yapılmış modellerin insan oylamalarında ön sıralarda yer aldığı görülmüştür. Bu durum, \textit{Türkçe-Kitap} veri kümesinin katkısıyla mümkün olmuştur. Aynı BDM'ye sahip modeller karşılaştırıldığında, LLaMA modelleri arasındaki varyasyonun düşük olduğu, performanslarının birbirine oldukça yakın olduğu görülmektedir. Buna karşın Mistral modelleri arasında puan açısından belirgin farklar mevcuttur. Eğitim kümeleri sabit tutulduğunda, hem ön eğitim hem de ince ayar kümesi aynı olduğunda, LLaMA modelleri genellikle üstün performans sergilemiştir. Ancak eğitimlerde $\text{CosmosVQA}_P$ + $\text{CosmosVQA}_F$ kümesi kullanıldığında bu durumun tam tersinin geçerli olduğu saptanmıştır. İnsan oylaması sonuçları Tablo \ref{table:human_voting_results}'te gösterilmiştir.

\subsection{İkili Sınıflandırıcı Analizi}
 Bu değerlendirme yöntemi modelin tek bir kelime ile sınıflandırma kabiliyetini ölçmektedir. Modelin kesinlik içeren ifadeler söylemesi beklenmektedir. Bu ifadeler kısa bir cümle olmalı veya tek bir kelimeden oluşmalıdır: "Evet." ya da "Hayır."
 Bu şartlara uyulmayan, uzun açıklamalar içeren model cevapları geçersiz sayılmıştır. Her bir sorunun tek bir doğru cevabı olduğu için modellerin başarıları "F1" ve "accuracy" gibi skorlarla ölçülmüştür. Ölçüm sonuçları Tablo \ref{table:model_metrics_table}'te bulunmaktadır.

Genel olarak bütün modellerin recall skorlarının 0.9'un üzerinde olduğu görülmüştür. Bu durum modellerin sorulara çoğu zaman "Evet." cevabını verdiğini göstermektedir.
Doğru cevabı "Evet." olmasına rağmen "Hayır." şeklinde sınıflandırılan örneklerin (yanlış negatif) az olması, bazı kritik görevlerde modellerin başarısını göstermektedir. 



Modeller için genelleyici olarak nitelendirilebilecek F1 skoruna bakıldığında en yüksek performansı LLaMA modellerinin sergilediği görülmüştür. Ancak mistral modelleri arasında \textit{Mistral + $\text{99eren}_P $ + $\text{99eren}_F$} modeli dışında daha düşük varyanslar saptanırken LLaMA modellerinde varyansın daha büyük olduğu da gözlemlenmiştir.

Her bir metrik, modellerin çeşitli özelliklerini farklı açılardan ortaya koymaktadır. İnsan oylaması sonuçları incelendiğinde, LLaMA tabanlı modeller arasındaki varyasyonun düşük olduğu, ancak ikili sınıflandırma sonuçlarında bu varyasyonun yüksek olduğu gözlemlenmiştir. Bu durumun tam tersi ise Mistral modelleri için geçerlidir. GPT-4o hakemliğinde elde edilen "Toplam" skoru ve insan oylamasından elde edilen elo skoru arasındaki korelasyon 0.948 çıkmıştır.  Hakem olarak kullanılan GPT-4o genel olarak insan hakemlere oldukça yakın kararlar vermiştir ama birbirine yakın skorları olan modeller arasındaki sıralamalarda insan hakemler ve GPT-4o arasında farklılıklar gözlenmiştir.

\section{Sonuçlar ve Gelecek Çalışmalar}
Bu çalışmada bir Türkçe görsel dil modeli geliştirilirken model mimarisi ve eğitim kümesi seçiminin performansa etkileri incelenmiştir. Denenen 8 kombinasyon arasında genel olarak yakın başarılar elde eden 4 model görülmektedir. Bu modellerin başarı sıralamaları çeşitli metriklerde farklılık göstermektedir. Hedeflenen görev, kullanılan mimari ve ön eğitim için tercih edilen veri kümesine göre optimum ince ayar kümesi değişiklik göstermiştir.

$\text{CosmosVQA}_F$ ile ince ayarı yapılmış modellerin başarıları, diğer modellere kıyasla yüksek olmuştur. \textit{LLaMA + $\text{CosmosVQA}_P$ + $\text{99eren}_F$} modeli hem ikili sınıflandırmada hem de GPT-4o hakemliğinde birinci olmuştur. Ayrıca GPT-4o'nun hakemliğine göre, \textit{LLaMA + $\text{99eren}_P$ + $\text{CosmosVQA}_F$} modelinin 97.7 puan alarak GPT4o-mini'yi OCR görevinde 1.5 puan farkla geçmesi dikkat çekicidir.

Bu çalışma Türkçe görsel dil modelleri alanında önemli bir ilk adım niteliğindedir. Eğitilen modeller ve elde edilen sonuçlar, Türkçede açık kaynaklı görsel dil modeli araştırmalarında gelecekteki çalışmalara referans olacaktır. Ayrıca bu çalışmanın farklı görsel senaryolara yönelik daha güçlü görsel dil modelleri geliştirilmesi için değerli bir başlangıç noktası olacağı düşünülmektedir.

Bu çalışmada kullanılan veri setleri açık kaynak olarak paylaşılmıştır \cite{llava-pretrain, llava-finetune, turkce-kitap}. Ayrıca, yüksek performans gösteren \textit{LLaMA + $\text{99eren}_P$ + $\text{CosmosVQA}_F$} modeli erişime sunulmuştur \cite{cosmosllava}. Bu modeli denemek için bir Hugging Face Space arayüzü hazırlanmıştır \cite{cosmosllava-space}.

Gelecek çalışmalarda modellerin genel performansını artırmak için veri çeşitliliği ile eğitim süreçlerinin genişletilip farklı görevler için özelleştirilmiş modeller üretilmesi  amaçlamaktadır.

\section{Teşekkür}
Bu çalışma Türkiye Bilimsel ve Teknolojik Araştırma Kurumu (TÜBİTAK) tarafından 124E055 numaralı proje ile desteklenmiştir.

\bibliographystyle{ieeetr} 
\bibliography{references} 

\end{document}